\documentclass{article}
\usepackage{spconf,amsmath,graphicx}

\usepackage{enumitem}
\setlist{nosep, leftmargin=14pt}

\usepackage{mwe} 
\usepackage{amssymb}
\usepackage[ruled,vlined]{algorithm2e}
\usepackage{booktabs}
\usepackage{multirow}
\usepackage{arydshln}
\usepackage{xcolor}
\usepackage{hyperref}

\newcommand{\red}[1]{\textcolor{black}{#1}}
\newcommand{\blue}[1]{\textcolor{black}{#1}}

\title{Hierarchical LoG Bayesian Neural Network for \\ Enhanced Aorta Segmentation}
\name{Delin An$^{1}$ \qquad Pan Du$^{2}$ \qquad Pengfei Gu$^{3}$ \qquad Jian-Xun Wang$^{2, 4}$ \qquad Chaoli Wang$^{1}$}

\address{$^{1}$ Department of Computer Science and Engineering, University of Notre Dame, Notre Dame, IN\\
    $^{2}$ Department of Aerospace and Mechanical Engineering, University of Notre Dame, Notre Dame, IN\\
    $^{3}$ Department of Computer Science, University of Texas Rio Grande Valley, Edinburg, TX\\
    $^{4}$ Sibley School of Mechanical and Aerospace Engineering, Cornell University, Ithaca, NY}
\begin{document}
%
\maketitle
\begin{abstract}
Accurate segmentation of the aorta and its associated arch branches is crucial for diagnosing aortic diseases. While deep learning techniques have significantly improved aorta segmentation, they remain challenging due to the intricate multiscale structure and the complexity of the surrounding tissues. This paper presents a novel approach for enhancing aorta segmentation using a Bayesian neural network-based hierarchical Laplacian of Gaussian (LoG) model. \red{Our model consists of a 3D U-Net stream and a hierarchical LoG stream: the former provides an initial aorta segmentation, and the latter enhances blood vessel detection across varying scales by learning suitable LoG kernels, enabling self-adaptive handling of different parts of the aorta vessels with significant scale differences.} We employ a Bayesian method to parameterize the LoG stream and provide confidence intervals for the segmentation results, ensuring robustness and reliability of the prediction for vascular medical image analysts. Experimental results show that our model can accurately segment main and supra-aortic vessels, yielding at least a 3\% gain in the Dice coefficient over state-of-the-art methods across multiple volumes drawn from two aorta datasets, and can provide reliable confidence intervals for different parts of the aorta. \blue{The code is available at \url{https://github.com/adlsn/LoGBNet}.}
\end{abstract}
%
%
\section{Introduction}
\label{sec:intro}

Semantic segmentation is a crucial research direction in deep learning technologies, extensively applied in medical image processing. It finds wide-ranging applications in organ segmentation, lesion detection, tumor segmentation, and vessel segmentation. Vessel segmentation includes coronary artery, aorta, and pulmonary artery segmentation. Among these, aorta segmentation presents unique challenges. While U-Net~\cite{Ronneberger2015} has successfully segmented other blood vessels, the aorta poses specific difficulties due to its characteristics. Other methods, like U-Net 3D~\cite{cciccek20163d}, are tailored for general abdominal organs but fall short of effectively segmenting the aorta and arch branches. Moreover, providing reliable confidence intervals for aorta segmentation results is challenging. Uncertainty quantification (UQ) offers analysts a basis for assessing the reliability of current predictions and significantly impacting downstream tasks involving the aorta. For instance, in computational fluid dynamics (CFD) simulations of the aorta, UQ helps verify and calibrate CFD models, ensuring they accurately reflect physiological conditions.

\red{The aorta is conformed by the aortic arch, ascending aorta, descending aorta, and numerous branches. These components exhibit significant variations in size and shape across individuals. Following~\cite{supra-aortic2023}, the main arch and branches of the aorta are referred to as the main aorta (MA) and supra-aorta branches (SA), respectively. Additionally, aorta segmentation requires higher computational power due to the need for high-resolution imaging.}
%
%
\red{Traditional aorta segmentation methods involve manual centerline tracking and boundary filtering~\cite{updegrove2017simvascular}. Deep learning models like U-Net reduce manual work, using encoders such as 3D CNNs, transformers, and attention-based models~\cite{oktay2018attention}. Some use labeled centerlines to improve performance~\cite{maher2021geometrica}, but obtaining them is challenging. Thus, automatic centerline extraction is preferred~\cite{zhao2021automatica}. These methods perform well on the MA but have limitations: they neglect the SA, and incorporating centerlines increases model complexity and training difficulty.}

\blue{We propose LoGB-Net, a Laplacian of Gaussian (LoG)-based Bayesian network, for aorta segmentation, enhancing multiscale vessel recognition (especially for SA) and improving model automation. LoGB-Net features two streams: a U-Net-based regular stream for overall aorta segmentation and a LoG stream with a Bayesian LoG module and balanced gate to refine local details.
Our key contributions include}
\begin{itemize}
  \item A hierarchical LoG module is introduced to improve the performance of aorta segmentation. This design enhances features of aortic vessels of various sizes in a self-adaptive manner, effectively emphasizing multiscale features and strengthening the model's segmentation capability.
  \item An imbalance problem between the aorta's foreground and background is addressed by designing a balanced gate in the LoG stream. This gate balances the ratio between multiscale blood vessels and prevents the model from overfitting, ensuring improved segmentation accuracy.
  \item The LoG module is parameterized to obtain the UQ for the segmentation results. This approach provides confidence intervals for the segmentation results, allowing for measuring the uncertainty in the model's predictions, reflecting the prediction's reliability in cases where the edges of the aorta are ambiguous, and offering UQ for downstream tasks in aortic analysis.
\end{itemize}

\section{Related Work}
\label{sec:rel}

Medical image semantic segmentation methods are semi-automatic and automatic~\cite{Jin2021}. Semi-automatic methods involve manual lumen boundary and centerline annotation, achieving good performance but are labor-intensive. SimVascular~\cite{updegrove2017simvascular} reduces manual effort by combining centerline and lumen edge for segmentation but still requires annotations for new data. To improve efficiency, machine learning methods, such as~\cite{vermeer2004model}, use Gaussian filters and Fourier transforms to identify regions of interest. However, they are noise-sensitive and require parameter tuning for different object sizes.

\red{Deep learning has shifted segmentation towards automation, with methods categorized based on centerline usage: with 
and without centerline. 
While centerlines improve performance, annotating them is labor-intensive, leading to a focus on the direct segmentation of raw images. Encoder-decoder models like U-Net, PSPNet~\cite{Zhao2017}, FC-DenseNet~\cite{Jegou2017}, and \blue{kCBAC-Net~\cite{GuZZWC21}} are widely used. U-Net excels in medical image processing tasks. Transformer-based models like UNETR~\cite{Hatamizadeh2022}, UNETR++~\cite{shaker2023unetr}, Swin-UNETR~\cite{swinunetr2022}, and \blue{ConvFormer~\cite{GuZWC23}} integrate attention mechanisms, performing well in blood vessel segmentation with consistent scale but struggle with the aorta and supra branches due to significant scale variation.
As a dual-stream model, Gated-SCNN~\cite{takikawa2019gatedscnn} combines a regular and a shape stream leveraging image gradients, achieving favorable results in small object detection. This model guides SA-enhancing architectures for aorta segmentation.
Furthermore, UQ provides confidence intervals for segmentation predictions, enhancing reliability in aortic research. Existing UQ methods often increase complexity by integrating VAE encoders into U-Net architectures. \cite{Sagar_2022_WACV} evaluated these approaches on brain tumor datasets.}

\begin{figure}[htb]
  \centering
  \includegraphics[width=1.0\linewidth]{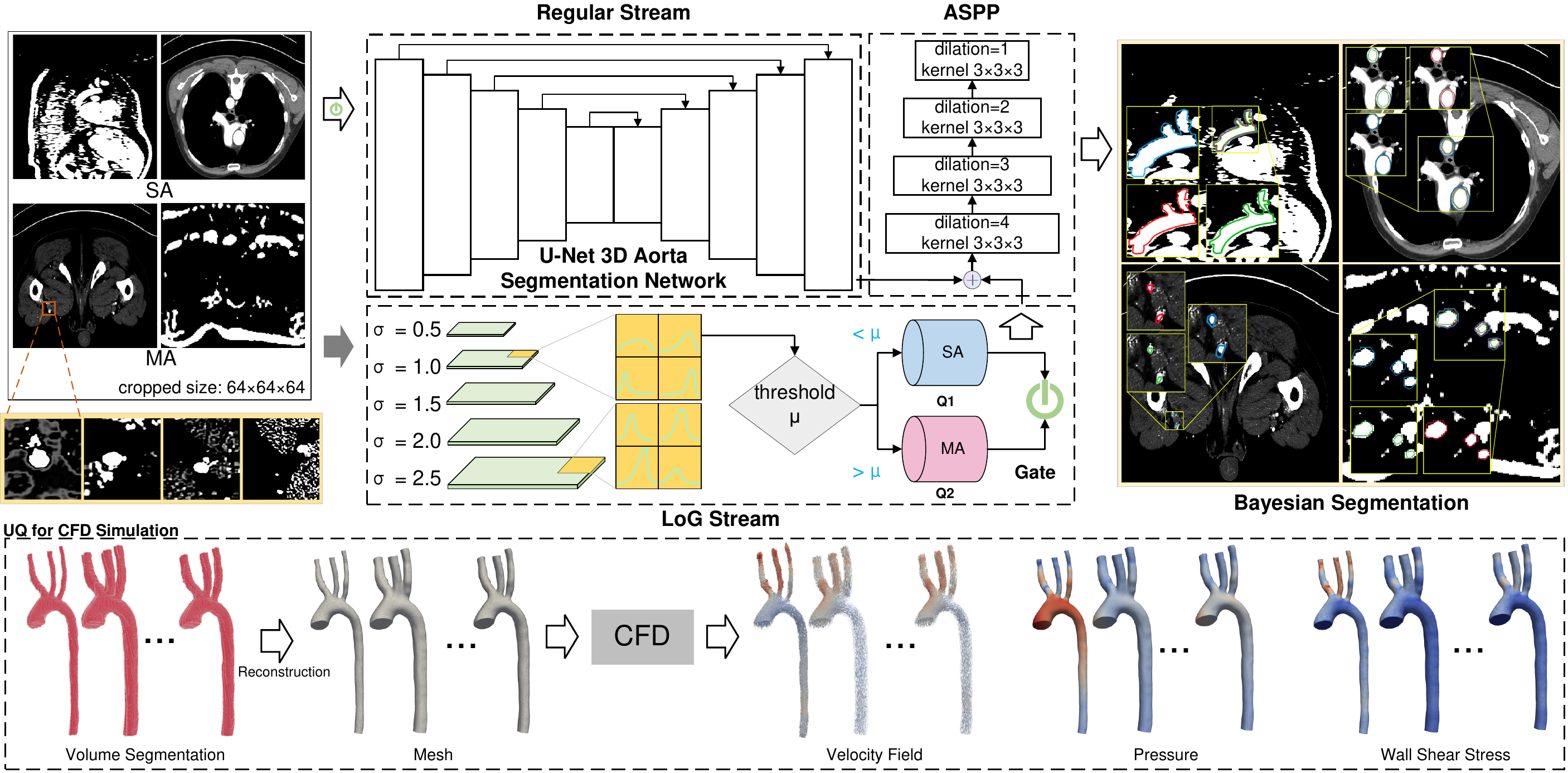} 
  \caption{The top shows the flow diagram of LoGB-Net. The Bayesian segmentation results' blue, red, and green lines represent the segmentation's upper bound, mean, and lower bound. The bottom displays the application of UQ in aorta segmentation for downstream tasks, such as CFD simulation.}
  \label{overview}
\end{figure}

\section{LoGB-Net}

Multiscale aortic vessels require attention to the overall structure and local details, and the imbalanced foreground-background ratio complicates training. We propose a hierarchical LoG module to address these issues and enhance aorta segmentation. Our framework (Figure~\ref{overview}) comprises regular and LoG streams. 
The regular stream employs a 3D U-Net~\cite{cciccek20163d} to segment the overall aorta structure. The LoG stream consists of a Bayesian hierarchical LoG module and a balanced gate, augmenting the segmentation with local details.
The balanced gate contains two queues $Q_1$ and $Q_2$ to store multiscale vessel images. When these queues are full, the gate will open for the regular and LoG streams. 

{\bf Hierarchical LoG module.} 
Encoder-decoder models effectively segment vessels with manageable complexity~\cite{shaker2023unetr,moccia2018blood}, achieving high-quality results for retinal~\cite{Yang2022} and liver~\cite{zeng2018automatic} vessels. However, further enhancement is required for the SA of the aorta to achieve satisfactory segmentation results. The LoG stream can strengthen aorta features by self-adapting to multiscale vessel characteristics.

The LoG model, \red{used by}~\cite{moccia2018blood}, is a second-order derivative operator for edge detection in images. Specifically, the 2D LoG model is defined as $ \nabla^2G(x,y) = \frac{\partial^2 G(x,y)}{\partial x^2} + \frac{\partial^2 G(x,y)}{\partial y^2}, \label{laplacian} $ where $G(x,y)$ is the Gaussian function: $ G(x,y) = \frac{1}{2\pi \sigma^2}e^{-\frac{x^2 + y^2}{2\sigma^2}}. \label{gaussian} $
\red{Since the Laplacian is sensitive to noise, applying Gaussian smoothing is necessary.} In \red{3D CTA images}, we adopt the 3D LoG model, defined as $ \nabla^2G(x,y,z) = \frac{\partial^2 G(x,y,z)}{\partial x^2} + \frac{\partial^2 G(x,y,z)}{\partial y^2} + \frac{\partial^2 G(x,y,z)}{\partial z^2}. $
\red{This continuous function is transformed into a 3D convolutional kernel through discretization. The standard deviation $\sigma$ and kernel size influence the detection size of target objects.} \cite{vermeer2004model} tested various parameters to fit retinal vessels, increasing manual operations. 

To leverage the LoG model's self-adaptive capability \red{for main and supra aorta detection}, we design a hierarchical LoG stream, as shown in Figure~\ref{overview}. This module comprises five 3D convolutional layers with kernel sizes \red{3, 5, 7, 9, and 11}, initialized with 3D LoG kernels at different \red{trainable} $\sigma$ values \red{(0.5 to 2.5)}. \red{Unlike the original LoG method with fixed parameters, we make them trainable in our approach.} The initial $\sigma$ values are optimized to achieve the best lumen boundaries. The module's multiple layers facilitate the self-adaptive enhancement of multiscale aortic vessels.

\begin{table*}[htbp]
  \caption{Quantitative results (reported as mean $\pm$ standard deviation) of different methods for the testing set. The best results are highlighted in bold. The experiment was conducted with varying random seeds five times to evaluate the outcomes.}
  \centering
  \resizebox{\textwidth}{!}{
    \begin{tabular}{ccccccccccccc}
      \toprule
      \multirow{2}{*}{}                 & \multicolumn{1}{c}{Type} & \multicolumn{4}{c}{CNN-based} & \multicolumn{6}{c}{Attention-based} & \multicolumn{1}{c}{Ours}                                                                                                                                                                                                       \\
      \cmidrule(lr){3-6} \cmidrule(lr){7-12} \cmidrule(lr){13-13}
                                        & Metric                   & FPN                           & U-Net 3D                            & PSPNet                   & nnUNet          & Attention-UNet  & MISSFormer                        & Swin-UNETR      & TransUNet       & UNETR           & UNETR++                           & LoGB-Net                          \\
      \midrule
      \multirow{3}*{\rotatebox{90}{SA}} & Dice $\uparrow$          & 0.722$\pm$0.023               & 0.764$\pm$0.042                     & 0.775$\pm$0.037          & 0.766$\pm$0.068 & 0.824$\pm$0.025 & 0.882$\pm$0.021                   & 0.896$\pm$0.019 & 0.823$\pm$0.022 & 0.878$\pm$0.031 & 0.892$\pm$0.037                   & \textbf{0.927}$\pm$\textbf{0.011} \\
                                        & ASD $\downarrow$         & 1.523$\pm$0.323               & 1.376$\pm$0.311                     & 1.329$\pm$0.304          & 1.214$\pm$0.322 & 0.929$\pm$0.213 & 0.813$\pm$0.196                   & 0.786$\pm$0.249 & 0.928$\pm$0.284 & 0.702$\pm$0.167 & 0.682$\pm$0.156                   & \textbf{0.678}$\pm$\textbf{0.168} \\
                                        & Hausdorff $\downarrow$   & 8.490$\pm$3.252               & 8.323$\pm$4.022                     & 9.765$\pm$3.773          & 8.941$\pm$2.547 & 6.725$\pm$0.294 & 6.899$\pm$0.077                   & 6.324$\pm$0.133 & 6.452$\pm$0.236 & 6.139$\pm$0.698 & \textbf{6.034}$\pm$\textbf{0.356} & 6.225$\pm$0.957                   \\
      \hdashline
      \multirow{3}*{\rotatebox{90}{MA}} & Dice $\uparrow$          & 0.734$\pm$0.016               & 0.773$\pm$0.026                     & 0.778$\pm$0.034          & 0.775$\pm$0.028 & 0.847$\pm$0.021 & 0.899$\pm$0.016                   & 0.909$\pm$0.033 & 0.904$\pm$0.023 & 0.904$\pm$0.036 & 0.907$\pm$0.023                   & \textbf{0.937}$\pm$\textbf{0.006} \\
                                        & ASD $\downarrow$         & 1.454$\pm$0.364               & 1.342$\pm$0.344                     & 1.335$\pm$0.313          & 1.397$\pm$0.311 & 0.879$\pm$0.225 & 0.923$\pm$0.233                   & 0.804$\pm$0.218 & 0.931$\pm$0.314 & 0.688$\pm$0.199 & 0.744$\pm$0.194                   & \textbf{0.682}$\pm$\textbf{0.197} \\
                                        & Hausdorff $\downarrow$   & 10.245$\pm$3.779              & 11.556$\pm$3.929                    & 9.987$\pm$4.201          & 9.294$\pm$2.326 & 6.942$\pm$0.309 & \textbf{6.125}$\pm$\textbf{0.191} & 7.021$\pm$0.065 & 6.878$\pm$0.254 & 6.422$\pm$0.453 & 6.321$\pm$0.241                   & 6.322$\pm$1.485                   \\
      \bottomrule
    \end{tabular}}
  \label{tab:quantitative1}
\end{table*}

{\bf Bayesian method and balanced gate.} 
\red{The low resolution of CTA images makes lumen edge segmentation challenging due to blurry boundaries. We address this by using a Bayesian method~\cite{Thiagarajan2022} to parameterize the LoG module, treating the LoG kernel as the prior distribution and the module's output as the posterior, defined as:} $ p(\theta|D) = \frac{p(D|\theta)p(\theta)}{p(D)}, $ \red{where $\theta$ are the LoG module parameters and $D$ is the image.}
\red{The optimization aims to maximize the Evidence Lower Bound (ELBO):} $ \theta=\mathrm{argmax}(\mathrm{ELBO}). $
\red{Our overall loss function combines Dice loss and ELBO loss:} $ \mathbb{L} = \mathrm{Dice} - \mathrm{ELBO}. \label{overall_loss} $ \red{We decrease $\mathbb{L}$ to maximize the ELBO value.}
We compared the LoG module with and without Bayesian parameterization in the ablation study. Our results demonstrate that the Bayesian parameterized LoG module produces more stable predictions.

Since the aorta occupies only a small section of the CTA image, there is an imbalance between the foreground (aorta) and the background. \red{To address this, we propose a balanced gate to achieve foreground-background balance, defined as:} $ T = \frac{\sum_{i=1}^{N_F}f_i}{\sum_{j=1}^{N_B}b_j}, $ \red{where $T$ denotes the foreground-to-background ratio. In Figure~\ref{overview}, the threshold $\mu$ is determined statistically by calculating the mean of the foreground-to-background ratio for the SA. Then, we introduce two counters, $c_1$ and $c_2$, associated with two queues, $Q_1$ and $Q_2$, each with a capacity of $C$. When the foreground area occupied by blood vessels is small, indicated by $T < \mu$ and $c_1 < C$, we enqueue the input image into $Q_1$. Conversely, when the foreground area is large, indicated by $T > \mu$ and $c_2 < C$, we enqueue the input image into $Q_2$. Once both $c_1$ and $c_2$ reach the capacity $C$, we concatenate the $2C$ images along the channel dimension and initiate the regular stream. In our experiments, we set $\mu=0.15$. This balanced gate ensures a more balanced ratio between large and small foreground regions, enhancing the model's robustness.}

{\bf Algorithm Description.}
The overall pipeline follows the steps below. (1) Input the CTA image into the LoG stream and calculate the mean of the output from the five layers. (2) Balance the input data ratio using the balanced gate. When counters $c_1$ and $c_2$ reach $C$, concatenate the $2C$ images along the channel dimension. (3) \blue{The outputs from the regular and LoG streams are concatenated and fed into a four-layer atrous spatial pyramid pooling (ASPP) module for further processing.} (4) Process the concatenated result using a 3D CNN with a $1\times1\times1$ kernel.

\section{Experiments}

{\bf Datasets, network training, and metrics.} 
Two datasets were used: the first is from~\cite{Wilson2013}, and the second is the Aortic Vessel Tree (AVT) CTA dataset~\cite{Radl2022}. The training set contains 34 volumes (24 from the first, 10 from the second), and the testing set has 16 volumes (8 from each).
Volumes were resampled using MONAI's {\sf Spacingd} function 
to $0.8\times0.8\times0.3$ spacing, cropped into $64\times64\times64$ blocks, and normalized to $[0,1]$. Data augmentation included {\sf RandFlipd} and {\sf RandRotate90d} (probability 0.1), applied only to training data.
We used the Adam optimizer. 
The hyperparameters are set as $\beta_1=0.8$ and $\beta_2=0.999$, while the penalty weight is set to $10^{-5}$. The learning rate is $10^{-4}$. The model is trained for 5000 epochs on an NVIDIA RTX 3080 GPU.
We adopted three metrics: Dice coefficient, average surface distance (ASD), and symmetric Hausdorff distance.

\begin{figure}[htbp]
  \centering
  \includegraphics[width=1.0\linewidth]{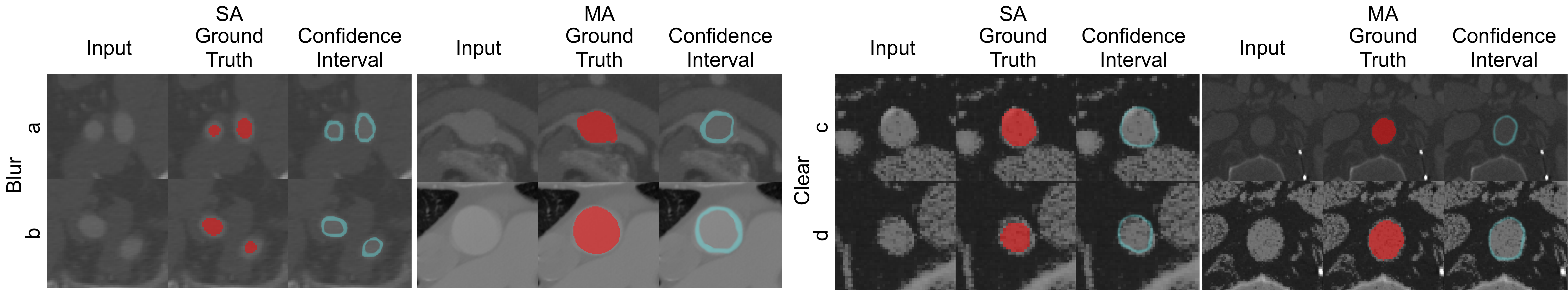} 
  \caption{Confidence intervals for aortic predictions based on boundary clarity variations.}
  \label{fig_uq}
\end{figure}

{\bf UQ.}
We assessed uncertainty due to varying boundary clarity and annotation errors by calculating confidence intervals from ten Bayesian inferences (Figure~\ref{fig_uq}). Blurred boundaries led to larger confidence intervals.

\begin{table*}[htbp]
  \caption{Quantitative results of the LoGB-Net's ablation study. $\mathcal{L}_{LoG}^-$, $\mathcal{L}_{Bay}^-$, and $\mathcal{L}_{Gate}^-$ represent LoGB-Net without LoG module, Bayesian optimization, and balanced gate, respectively. $\mathcal{L}(1)$, $\mathcal{L}(2)$, $\mathcal{L}(3)$, $\mathcal{L}(4)$, and $\mathcal{L}(5)$ represent LoGB-Net with 1, 2, 3, 4, and 5 LoG layers, respectively.}
  \centering
  \resizebox{\textwidth}{!}{
      \begin{tabular}{cccccccccc}
          \toprule
                                               & Metric                 & $\mathcal{L}_{LoG}^-$ ($\mathcal{L}(0)$) & $\mathcal{L}_{Bay}^-$ & $\mathcal{L}_{Gate}^-$ & $\mathcal{L}(1)$ & $\mathcal{L}(2)$ & $\mathcal{L}(3)$ & $\mathcal{L}(4)$ & $\mathcal{L}(5)$ (Ours)           \\
          \midrule
          \multirow{3}*{\rotatebox{90}{SA}} & Dice $\uparrow$        & 0.735$\pm$0.033                          & 0.863$\pm$0.031       & 0.896$\pm$0.012        & 0.852$\pm$0.033  & 0.863$\pm$0.018  & 0.872$\pm$0.011  & 0.893$\pm$0.033  & \textbf{0.927}$\pm$\textbf{0.011} \\
                                               & ASD $\downarrow$       & 1.384$\pm$0.334                          & 0.810$\pm$0.214       & 0.801$\pm$0.179        & 1.291$\pm$0.123  & 1.226$\pm$0.202  & 0.922$\pm$0.244  & 0.772$\pm$0.202  & \textbf{0.678}$\pm$\textbf{0.168} \\
                                               & Hausdorff $\downarrow$ & 8.271$\pm$2.276                          & 7.993$\pm$2.112       & 7.909$\pm$1.093        & 8.203$\pm$1.903  & 8.014$\pm$1.221  & 7.027$\pm$0.721  & 6.626$\pm$0.121  & \textbf{6.225}$\pm$\textbf{0.957} \\
          \hdashline
          \multirow{3}*{\rotatebox{90}{MA}} & Dice $\uparrow$        & 0.741$\pm$0.022                          & 0.871$\pm$0.033       & 0.882$\pm$0.021        & 0.866$\pm$0.014  & 0.865$\pm$0.009  & 0.877$\pm$0.036  & 0.923$\pm$0.003  & \textbf{0.937}$\pm$\textbf{0.006} \\
                                               & ASD $\downarrow$       & 1.399$\pm$0.123                          & 0.961$\pm$0.120       & 0.933$\pm$0.200        & 1.105$\pm$0.239  & 0.979$\pm$0.103  & 0.933$\pm$0.210  & 0.727$\pm$0.121  & \textbf{0.682}$\pm$\textbf{0.197} \\
                                               & Hausdorff $\downarrow$ & 8.335$\pm$2.996                          & 8.220$\pm$2.003       & 8.011$\pm$1.898        & 8.277$\pm$2.516  & 8.100$\pm$2.274  & 7.965$\pm$1.001  & 7.482$\pm$1.112  & \textbf{6.322}$\pm$\textbf{1.485} \\
          \bottomrule
      \end{tabular}}
  \label{tab:quantitative2}
\end{table*}

{\bf Evaluation results.} 
We compared LoGB-Net with CNN-based (FPN~\cite{Lin_2017_CVPR}, U-Net 3D~\cite{cciccek20163d}, PSPNet~\cite{Zhao2017}, nnUNet~\cite{isensee2021nnu}) and attention-based (Attention-UNet~\cite{oktay2018attention}, MISSFormer~\cite{huang2021missformer}, Swin-UNETR~\cite{swinunetr2022}, TransUNet~\cite{Chen2021}, UNETR~\cite{Hatamizadeh2022}, UNETR++~\cite{shaker2023unetr}) methods. LoGB-Net's computational cost is 48.88 GFLOPS, modestly above UNETR's 41.19 GFLOPS but significantly lower than nnUNet's 412.65 GFLOPS. Using the same training and data augmentation for all models, Table~\ref{tab:quantitative1} shows that LoGB-Net outperforms all baselines across various metrics, particularly achieving approximately 3\% higher Dice coefficient than Swin-UNETR.

\begin{figure}[htbp]
  \centering
  \includegraphics[width=1.0\linewidth]{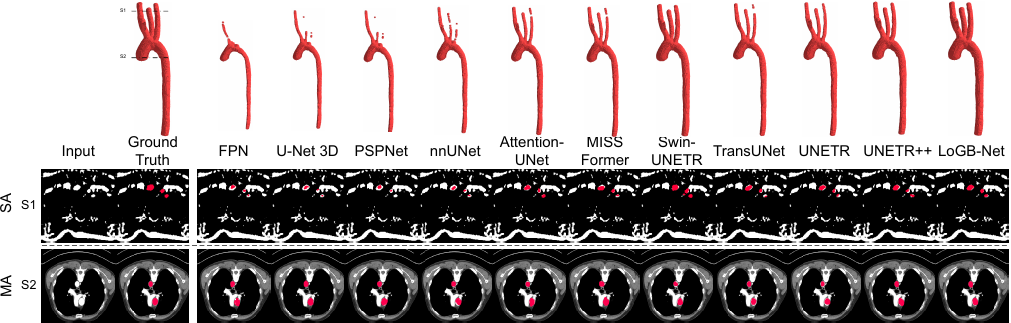} 
  \caption{Qualitative results of different methods for one volume in the testing set (drawn from the first dataset). The top row displays the 3D segmentation prediction, where selected slices (S1 to S2) are marked on the ground-truth result.}
  \label{fig2}
\end{figure}

As illustrated in Figure~\ref{fig2}, while most methods perform similarly on MA vessels, LoGB-Net outperforms others on SA vessels, detecting small-radius vessels and producing smoother boundaries. The Bayesian segmentation results (Figure~\ref{overview}) further demonstrate LoGB-Net's robustness, showing minimal deviation in the Dice metric (Table~\ref{tab:quantitative1}).

\begin{figure}[htb]
  \centering
  \includegraphics[width=1.0\linewidth]{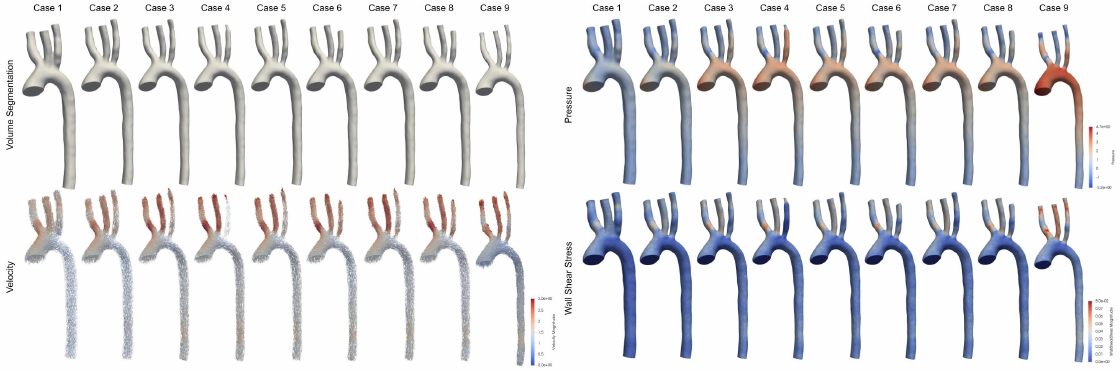} 
  \caption{CFD simulation results for the aortic vessel's velocity, pressure, and wall shear stress.
    Each case represents a different UQ realization.}
  \label{fig:CFD}
\end{figure}

\begin{figure}[htb]
  \centering
  \includegraphics[width=1.0\linewidth]{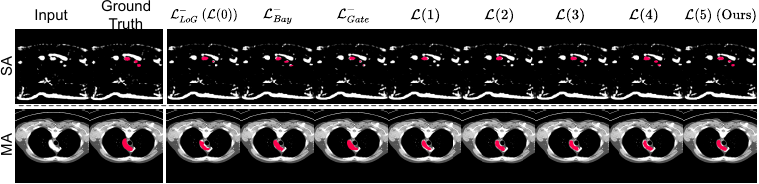} 
  \caption{Qualitative results of the LoGB-Net's ablation study.}
  \label{fig:ablation}
\end{figure}

{\bf CFD simulation.} 
We evaluated segmentation effectiveness by conducting CFD simulations on nine aorta shapes sampled from the posterior distribution. Using vmtk~\cite{VMTK} for mesh generation and OpenFOAM~\cite{Openfoam} to solve the Navier-Stokes equations, Figure~\ref{fig:CFD} compares simulation results among different UQ realizations. Results show significant flow variations, demonstrating sensitivity to segmented geometry and underscoring our model's UQ capability.

{\bf Ablation study.}
An ablation study assessed each component's effectiveness. Table~\ref{tab:quantitative2} shows that the LoG module, Bayesian optimization, and balanced gate each contribute to performance enhancement. Figure~\ref{fig:ablation} presents the qualitative results of the ablation study.

\section{Conclusions and Future Work}

\red{We have presented LoGB-Net, a hierarchical, Bayesian-optimized LoG segmentation model for enhanced aorta segmentation. Combining a regular stream for overall structure with a LoG stream for local details, LoGB-Net enhances SA detection, addresses foreground-background imbalance, and improves robustness by providing the UQ confidence intervals, outperforming baseline methods as demonstrated in experiments. The hierarchical structure and trainable parameters enable the detection of vessels with significant scale differences, particularly for SA. We will assess generalizability for future work by applying our method to other blood vessel datasets beyond the aorta.}

\section{Compliance with ethical standards}

This research study was conducted retrospectively using human subject data made available in open access by two publicly available datasets~\cite{Wilson2013,Radl2022}. Ethical approval was not required, as confirmed by the licenses attached to the open-access datasets.

\section{Acknowledgments}
This research was supported in part by NSF grants IIS-1955395, OAC-2047127, IIS-2101696, OAC-2104158, and IIS-2401144, and NIH grant 1R01HL177814-01. The authors would like to thank the anonymous reviewers for their insightful comments.

\bibliographystyle{IEEEbib}
\small{
\bibliography{refs-abbv}}

\end{document}